\documentclass[letterpaper]{article} 
\usepackage{aaai2027} 
\usepackage[hyphens]{url} 
\usepackage{graphicx} 
\usepackage{natbib} 
\usepackage{caption} 
\usepackage{booktabs}
\usepackage{colortbl}
\usepackage{dblfloatfix}
\usepackage{amsmath}
\usepackage{amssymb}
\usepackage{algorithm}
\usepackage{algpseudocode}

\nocopyright

\title{From Intent to Evidence: Policy-Steered Multi-Strategy Retrieval for Long-Video Agents}
\author{Can Zhang\textsuperscript{\rm 1,3}, Baofeng Zhang\textsuperscript{\rm 2,3}, Xiaotian Han\textsuperscript{\rm 3}\thanks{Project Lead.}, Junyuan Shang\textsuperscript{\rm 3}, Yuchen Ding\textsuperscript{\rm 3}, Shuohuan Wang\textsuperscript{\rm 3}, Dianhai Yu\textsuperscript{\rm 3}, Ruirui Li\textsuperscript{\rm 1}\thanks{Corresponding author.}}
\affiliations{
\textsuperscript{\rm 1}Beijing University of Chemical Technology\\
\textsuperscript{\rm 2}Peking University\\
\textsuperscript{\rm 3}Baidu, Inc.\\
{\tt\small alexlessend@gmail.com, boffinzhang@stu.pku.edu.cn, hanxiaotian@baidu.com, shangjunyuan@baidu.com, dingyuchen@baidu.com, wangshuohuan@baidu.com, yudianhai@baidu.com, ilydouble@gmail.com}
}

\begin{document}

\maketitle

\begin{abstract}
Existing long-video agents acquire evidence through one uniform behavior, ignoring whether the required evidence is concentrated, requires broad occurrence coverage, or must discriminate competing hypotheses---which can cause failure before substantive reasoning begins. Prescribing a fine-grained solution procedure for every question is not a satisfactory remedy, as it restricts autonomous exploration. We propose VESTA, a training-free long-video agent organized as a route-conditioned acquire--verify--consolidate loop. Before exploration, an intent router infers an evidence-acquisition policy---focused, recall, or contrastive retrieval over a shared visual--speech scene index---together with an evidence-accounting policy that configures the evidence view maintained during exploration. Policy-steered retrieval yields provisional references that multimodal evidence operations convert into observations, while the Reasoner remains free to verify them, re-query using intermediate findings, or inspect regions outside the retrieved set. A temporal evidence ledger consolidates observations into an adaptive, compressed view of temporal location, provenance, coverage, conflicts, verification outcomes, and hypothesis support, exposing missing and unresolved evidence to guide subsequent acquisition; finalization prioritizes verified observations. On Video-MME-v2, VESTA improves average accuracy by 2.7 points over VideoARM and gains across all six reported metrics. On LongVideoBench, EgoSchema, and LVBench under shared query-time models, it improves by 6.9 points on the LongVideoBench long subset and 1.5 on LVBench, and matches VideoARM on EgoSchema.
\end{abstract}

\section{Introduction}

As video increasingly serves as a primary medium for recording real-world activity and communicating knowledge, long-video understanding has become a fundamental capability for multimodal intelligence. Unlike short clips, the meaning of a long video often emerges from evolving events, interactions, and state changes whose relationships cannot be explained by any single local segment. A capable system must therefore recognize fine-grained visual and linguistic cues while connecting observations across extended temporal spans. Recent multimodal large language models (MLLMs) provide increasingly strong unified perception, cross-modal integration, and semantic reasoning capabilities, creating new opportunities for comprehensive long-video understanding \citep{team2023gemini,bai2025qwen2,wang2025internvideo2}.

Directly extending these capabilities to long videos, however, remains constrained by input scale. As duration grows, visual events, speech, and on-screen text accumulate continuously, making exhaustive high-resolution processing prohibitively expensive \citep{shen2024longvu}. More importantly, evidence relevant to a particular question usually occupies only a small fraction of the timeline and may appear sparsely, recur at multiple moments, or be distributed across distant intervals. The central challenge is therefore not only whether a model can reason from evidence once it is available, but whether a system can first locate and acquire the required evidence under a limited computational budget. Effective evidence acquisition is thus a prerequisite for subsequent understanding and reasoning.

Early LLM/MLLM-based approaches address this challenge through predefined pipelines that connect video processing, evidence selection, and language reasoning. LLoVi converts local clips into textual descriptions before answering through a prescribed aggregation procedure \citep{zhang2024simple}, while VideoTree narrows the analysis through a fixed sequence of clustering, relevance estimation, and hierarchical expansion \citep{wang2025videotree}. These designs reduce the amount of video content exposed to the language model, but they also prescribe how evidence should be organized and how reasoning should proceed, limiting the ability of stronger foundation models to use tools autonomously, explore the video, and adapt their strategy in response to intermediate observations.

As MLLMs have gained stronger tool-use and decision-making abilities, video agents have begun to acquire evidence actively through external memories and tool calls. Memory-augmented VideoAgent constructs temporal and object memories that an LLM queries interactively \citep{fan2024videoagent}; OmAgent combines a retrievable video store with task decomposition and tool use \citep{zhang2024omagent}; and DVD enables autonomous search over a multi-granularity video database \citep{zhang2025deep}. Compared with fixed pipelines, these systems preserve substantially more freedom for the model to decide where and how to investigate. Yet they generally rely on a uniform memory-access or search behavior across questions, without explicitly distinguishing whether the required evidence is concentrated, whether broader temporal coverage is necessary, or whether evidence must be organized around competing hypotheses. Consequently, an otherwise capable agent may fail before substantive reasoning begins: its initial candidates may collapse onto one temporal region, omit relevant occurrences, or provide imbalanced evidence for competing answers.

This limitation creates a design tension. Evidence acquisition needs structure, but replacing a uniform behavior with a fine-grained, tutorial-style procedure for every question is not the answer. Such procedures would prescribe how reasoning should unfold and reduce the agent's ability to revise its search after intermediate observations, form new hypotheses, or inspect unanticipated regions. The desired guidance should therefore operate at the level of \emph{what evidence must be acquired}, while leaving \emph{how the agent reasons and explores afterward} open.

To resolve this tension, we propose \textbf{VESTA} (\textbf{V}ideo \textbf{E}vidence \textbf{S}teering and \textbf{T}rust \textbf{A}ccounting), a training-free long-video agent for policy-driven evidence acquisition and verification. VESTA organizes exploration as a \emph{route-conditioned acquire--verify--consolidate} loop. Before exploration, VESTA invokes the intent router once over the question and optional candidate answers to infer an evidence-acquisition policy and an evidence-accounting policy. The former selects \emph{focused} acquisition when relevant evidence is expected to be concentrated, \emph{recall} acquisition when broad coverage over potentially recurring evidence should be prioritized, or \emph{contrastive} acquisition when competing hypotheses must be distinguished; the latter specifies which evidence views the Reasoner should track. These policies structure what evidence is acquired and consolidated, not the subsequent reasoning steps, and remain independent of question difficulty.

Conditioned on the routed policy, VESTA combines policy-steered retrieval with multimodal evidence operations inside a unified exploration loop. Every retrieved region remains a provisional reference: the Reasoner may inspect it visually, examine its speech, analyze a local clip against a targeted subquestion, re-query the scene index using intermediate observations, or explore intervals never returned by the retriever. As observations accumulate, the temporal evidence ledger consolidates them into an adaptive, compressed view that makes temporal location, provenance, coverage, conflicts, and verification outcomes explicit. Missing or unresolved evidence exposed by this view drives the next acquisition step, while finalization prioritizes verified observations. VESTA thereby provides question-matched structure at the evidence-acquisition level while preserving autonomous reasoning, exploration, and self-correction. Figure~\ref{fig:case-study} illustrates how this design combines cross-scene comparison with speech and local visual verification to recover a narrative transition missed by a uniform exploration baseline.

We evaluate VESTA on four benchmarks with complementary temporal regimes. On Video-MME-v2 \citep{fu2026videommev2}, VESTA improves overall accuracy by 2.7 points over VideoARM (50.3\% vs.\ 47.6\%) with gains across all six reported metrics. On the LongVideoBench long subset \citep{wu2024longvideobench}, EgoSchema \citep{mangalam2023egoschema}, and LVBench \citep{wang2025lvbench} under shared query-time models, the margin scales with the temporal span over which evidence must be located: +6.9 on 15--60 minute videos, +1.5 on hour-long videos, and parity on three-minute clips, where long-range search is not the bottleneck.

Our contributions are threefold:
\begin{itemize}
    \item We identify a structural mismatch in long-video agents: capable agents commonly use a uniform evidence-acquisition behavior despite different requirements for evidence concentration, comprehensive coverage, and hypothesis discrimination. We further articulate why tutorial-style workflows are not a satisfactory remedy, as they trade this mismatch for reduced exploration autonomy.
    \item We propose VESTA, a training-free long-video agent organized as a route-conditioned acquire--verify--consolidate loop. Its intent router selects focused, recall, or contrastive evidence acquisition without prescribing the Reasoner's subsequent tool use, hypothesis formation, or exploration beyond the initial candidates.
    \item We introduce a temporal evidence ledger that consolidates accumulated observations into an adaptive, compressed view of provenance, temporal coverage, evidence relations, conflicts, and hypothesis support. The ledger exposes unresolved evidence to guide subsequent acquisition and presents verification outcomes that inform finalization. VESTA improves over VideoARM across all six Video-MME-v2 metrics and across four benchmarks that span 3-minute to hour-long videos.
\end{itemize}
\begin{figure*}[!htbp]
    \centering
    \includegraphics[width=\textwidth]{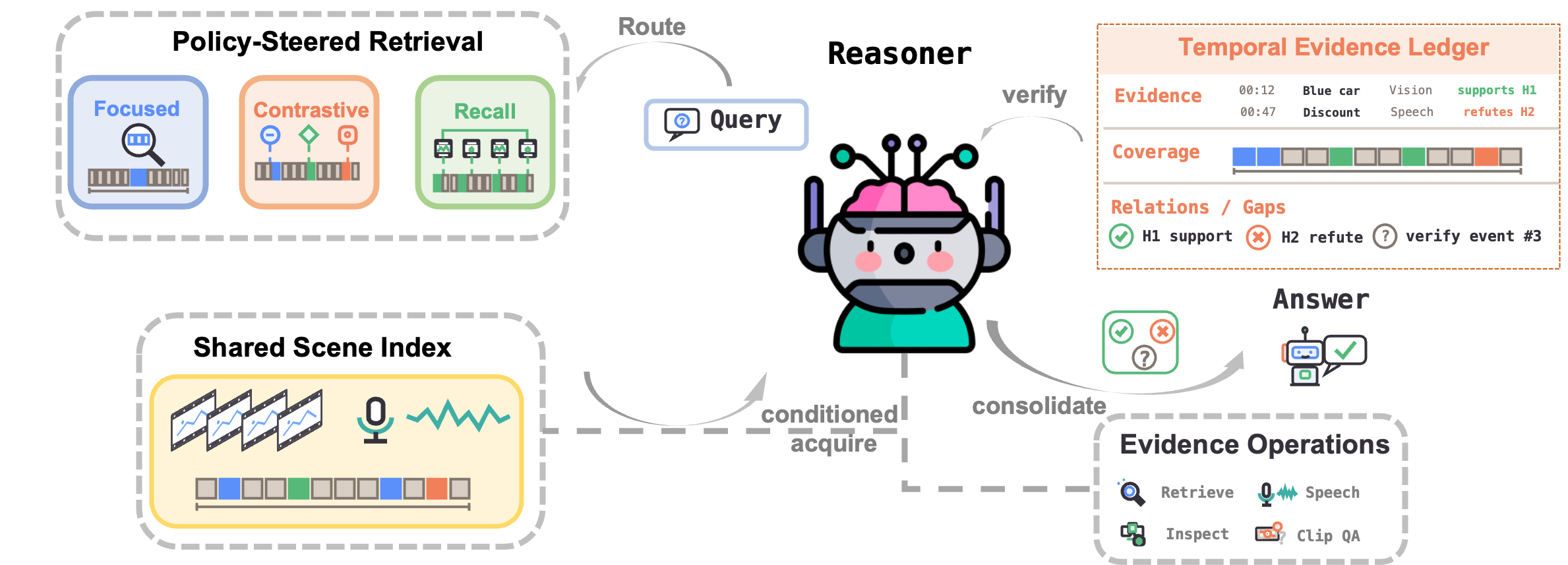}
    \caption{Overview of VESTA's route-conditioned acquire--verify--consolidate loop. The intent router selects focused, recall, or contrastive evidence acquisition over a shared visual--speech scene index. Retrieved scenes enter the Reasoner as provisional references, while shared evidence operations support autonomous re-retrieval and multimodal verification. The temporal evidence ledger consolidates observations into explicit evidence, coverage, and relation/gap views that guide subsequent exploration and inform answer finalization.}
    \label{fig:framework}
\end{figure*}

\section{Related Work}

\paragraph{MLLMs for Long-Video Understanding.}
Model-centric approaches extend multimodal backbones to process longer videos, including LLaVA-Video-7B \citep{zhang2025llavavideo}, Video-LLaMA \citep{zhang2023videollama}, LongVU \citep{shen2024longvu}, VideoChat-Flash \citep{li2024videochat}, and InternVideo2.5 \citep{wang2025internvideo2}. They expand visual context windows, sample more frames uniformly or sparsely, compress visual tokens, or integrate long-range information inside the model \citep{team2023gemini,wei2025visual,zhang2024long,yang2025pvc,tao2025dycoke}. Their ability to preserve evidence is nevertheless bounded by the visual-token and computational budgets available to the backbone: processing cost grows with video length, while aggressive compression can discard the fine-grained cues on which an answer depends. More fundamentally, representation expansion and compression generally optimize what the backbone can ingest, rather than decide whether a particular question requires concentrated evidence, comprehensive occurrence coverage, or hypothesis-discriminative evidence. VESTA does not modify the video backbone or increase its context budget; it determines what evidence should be acquired before that evidence is passed to the Reasoner.

\paragraph{Video Agents.}
Video agents combine ReAct-style control, task decomposition, and multi-round prompting with external evidence substrates, allowing an MLLM to decide what to inspect next, invoke tools, and revise its trajectory from intermediate observations \citep{yao2023react,yang2025vca,yin2026videoarm,liu2026longvideoagent}. Their evidence substrates include video databases and retrieval systems such as DVD \citep{zhang2025deep} and VideoRAG \citep{jeong2025videorag}, as well as structured memories such as VideoAgent \citep{fan2024videoagent}, MA-LMM \citep{he2024malmm}, Video-EM \citep{wang2025videoem}, WorldMM \citep{yeo2026worldmm}, and VideoLucy \citep{zuo2025videolucy}. These methods show that strong MLLMs can autonomously decompose questions, invoke tools, access external evidence, and revise their exploration trajectories. The remaining limitation is not a lack of reasoning or exploration capability: prompted loops commonly retain one general evidence-acquisition behavior across questions, while external stores primarily determine \emph{where evidence is kept}, not \emph{how different questions should acquire it}. Moreover, accumulative memories do not necessarily distinguish provisional retrievals from verified evidence or explicitly represent coverage gaps and conflicts. Prescribing a detailed workflow for each question would address this mismatch only by reducing exploration autonomy. VESTA instead intervenes at a coarser level: it makes access to a shared scene index question-conditioned without prescribing how the Reasoner must solve the question, and consolidates heterogeneous observations into a temporal evidence ledger that exposes coverage, conflicts, and unresolved evidence.

\section{Method}

\subsection{Overview}

Given a long video $V$, a question $Q$, and an optional candidate-answer set $C$, VESTA aims to produce an answer $A$ by actively acquiring and integrating relevant video evidence within an interaction budget $N$. Rather than prescribing fine-grained solution procedures, VESTA provides strategy-level guidance over evidence acquisition while leaving the Reasoner free to adapt its reasoning, tool use, and exploration to intermediate observations.

VESTA organizes the complete evidence exploration process as a \emph{route-conditioned acquire--verify--consolidate} loop. Before exploration, an intent router infers the question's evidence-acquisition requirement and selects an evidence-acquisition policy. Conditioned on this policy, policy-steered retrieval produces provisional references, and multimodal evidence operations convert selected references into observations; the Reasoner verifies, supplements, or corrects these observations, and the temporal evidence ledger adaptively consolidates them into a compressed, structured evidence view. By exposing evidence that remains uncovered or unresolved, this view drives subsequent acquisition and verification until the evidence supports an answer or the budget is exhausted. As illustrated in Figure~\ref{fig:framework}, this loop provides structured guidance over evidence acquisition while preserving autonomous reasoning, exploration, and self-correction.

\subsection{Intent Routing: From Question to Evidence-Acquisition Policy}
\label{sec:intent-routing}

Before exploring the video, VESTA invokes the intent router once over $Q$ and the optional candidate set $C$. The router neither accesses video content nor predicts the answer; it only determines the structure of evidence that subsequent exploration should acquire. Formally, it outputs a pair of policies
\begin{equation}
(\pi,\mu)=\mathcal{R}(Q,C),
\end{equation}
where $\pi$ is an \emph{evidence-acquisition policy} that selects how initial candidate evidence is obtained, and $\mu$ is an \emph{evidence-accounting policy} that specifies which evidence views the Reasoner should emphasize. The router's role ends after producing $(\pi,\mu)$: it does not record observations, update evidence state, or participate in verification.

The evidence-acquisition policy is
\begin{equation}
\pi \in \{\mathit{focused},\mathit{recall},\mathit{contrastive}\}.
\end{equation}
Focused acquisition prioritizes precision when relevant evidence is expected to be concentrated in a small number of temporal regions. Recall acquisition prioritizes coverage when a reliable answer benefits from recovering relevant occurrences across the timeline. Contrastive acquisition prioritizes comparability when the answer requires distinguishing competing hypotheses. These policies describe evidence requirements rather than question types or difficulty: concentrated evidence may require difficult reasoning, comprehensive coverage need not imply complex reasoning, and hypothesis discrimination does not imply that evidence must be temporally dispersed.

Unlike intent signals used only as prompt annotations or score weights, VESTA's routing result directly selects the retrieval strategy that is executed and configures the evidence view presented to the Reasoner. It changes how candidates enter the agent, not the Reasoner's available operations, search range, or reasoning procedure. All policies share one scene index, the same multimodal evidence operations, and a common interaction budget. The router requires no additional training or dataset-specific supervision.

\subsection{Policy-Steered Evidence Retrieval}
\label{sec:retrieval-strategies}

To provide a shared and reusable retrieval basis for all evidence-acquisition policies, VESTA first constructs a query-independent scene cache for each video. It partitions the video into scenes $\mathcal{S}=\{s_i\}_{i=1}^{M}$ using visual transitions subject to minimum and maximum duration constraints, and stores the global frame and temporal interval of every scene. For each $s_i$, VESTA samples a small set of representative frames and collects speech transcripts that overlap its interval. The scene encoder $f_s$ jointly encodes the visual frames and available speech into a scene representation, which is persisted with its temporal metadata in a shared vector index. Cache construction is query-independent, performed once per video, and reused by all questions, retrieval policies, and subsequent re-queries over that video.

For a text query $q$, the query encoder $f_q$ maps it into the same representation space, and the retriever scores scene $s_i$ as
\begin{equation}
r_i(q)=\operatorname{sim}\!\left(f_q(q),f_s(s_i)\right).
\end{equation}
The routed policy $\pi$ changes neither the scene cache, encoders, nor similarity function; it selects how queries are constructed and how the relevance ranking forms a candidate set $\mathcal{R}_{\pi}$. Thus, all three strategies share the same retrieval basis and downstream interface while producing differently structured initial evidence.

\paragraph{Focused retrieval.}
Focused retrieval provides a precision-oriented starting point when relevant evidence is likely concentrated in a small number of temporal regions. VESTA constructs a single query from $Q$ and returns the $K$ highest-scoring scenes:
\begin{equation}
\mathcal{R}_{\mathrm{focused}}
=\operatorname{TopK}_{s_i\in\mathcal{S}} r_i(Q).
\end{equation}
This provides a high-precision starting point without assuming that subsequent reasoning is easy or short.

\paragraph{Recall retrieval.}
Coverage-oriented questions expose a different failure mode: neighboring scenes from one high-similarity interval can monopolize standard top-$K$ results, leaving other relevant occurrences unretrieved. To broaden the initial evidence coverage, VESTA first obtains an enlarged pool
\begin{equation}
\mathcal{P}=\operatorname{Top}_{\alpha K,\,s_i\in\mathcal{S}}\,r_i(Q),
\qquad \alpha>1,
\end{equation}
and then greedily retains high-scoring scenes that are sufficiently separated in time. If fewer than $K$ scenes satisfy the separation condition, the remaining highest-scoring candidates backfill the set. The resulting $\mathcal{R}_{\mathrm{recall}}$ preserves semantic relevance while broadening temporal coverage over more potential occurrences. This is a bounded expansion over the same index, not an unbounded scan of the video, and it does not guarantee recovery of every relevant occurrence.

\paragraph{Contrastive retrieval.}
A third acquisition requirement arises when the answer depends on distinguishing competing hypotheses $\mathcal{H}=\{h_j\}_{j=1}^{J}$. Because a single query can bias retrieval toward one lexically salient hypothesis, VESTA forms $q_j=(Q,h_j)$ for each hypothesis, retrieves a small candidate set for each, and merges the results by temporal location:
\begin{equation}
\begin{aligned}
\mathcal{R}_{\mathrm{contrastive}}
&=\operatorname{Dedup}_{\mathrm{time}}\!\left(
\bigcup_{j=1}^{J}\operatorname{Top}_{k_j} r_i(q_j)
\right), \\
\sum_{j=1}^{J} k_j &\approx K.
\end{aligned}
\end{equation}
Each hypothesis therefore receives evidence that can support or refute it, while the total candidate scale remains comparable across policies.

Every strategy outputs temporally localized, scored scenes as \emph{provisional references}. These strategies alter the structure of initial candidate evidence, not the range that the Reasoner may subsequently explore. Retrieval results are not verified evidence and cannot prevent re-retrieval or inspection outside $\mathcal{R}_{\pi}$.

\subsection{Agentic Evidence Exploration and Verification}
\label{sec:exploration}

After policy-steered retrieval produces initial candidates, VESTA enters a multi-round evidence exploration process driven by the Reasoner. At step $t$, its decision context is
\begin{equation}
H_t=(Q,\pi,\mathcal{R}_{\pi},\mathcal{L}_t,\mathcal{M}_t),
\end{equation}
where $\mathcal{R}_{\pi}$ comprises the initial candidates together with any references added by mid-exploration re-queries, $\mathcal{M}_t$ is the accumulated structured tool memory containing observation details, and $\mathcal{L}_t$ is the compact temporal evidence ledger that organizes those observations by relevance, coverage, conflicts, and verification outcomes. The Reasoner selects the operation that best addresses the current evidence gap:
\begin{equation}
\begin{aligned}
a_t &= \mathcal{D}(H_t), \qquad a_t\in\mathcal{A}, \\
\mathcal{A}
&=\{a^{\mathrm{ret}},a^{\mathrm{inspect}},a^{\mathrm{speech}}, \\
&\quad a^{\mathrm{clip}},a^{\mathrm{final}}\}.
\end{aligned}
\end{equation}
The acquisition policy supplies a starting structure but never replaces this dynamic decision process.

All policies expose the same complementary evidence operations. \emph{Semantic Scene Retrieval} $a^{\mathrm{ret}}$ searches the shared scene index and allows refined queries based on newly observed information. \emph{Temporal Visual Inspection} $a^{\mathrm{inspect}}$ samples globally indexed frames from selected intervals to obtain coarse visual observations. \emph{Temporally Grounded Speech Transcription} $a^{\mathrm{speech}}$ transcribes speech within specified intervals and aligns it to the global timeline. \emph{Question-Conditioned Clip Analysis} $a^{\mathrm{clip}}$ asks a targeted subquestion over a local clip and returns a focused visual judgment with confidence information. \emph{Answer Finalization} $a^{\mathrm{final}}$ terminates exploration once the evidence state supports an answer.

Initial and mid-exploration retrievals remain unverified references rather than search-space constraints. Executing $a_t$ produces an observation $o_t$ with temporal and provenance information, after which VESTA updates the ledger:
\begin{equation}
\mathcal{L}_{t+1}=\mathcal{U}(\mathcal{L}_t,o_t,a_t).
\end{equation}
The updated ledger returns verification outcomes and unmet evidence requirements to the Reasoner, which may inspect a candidate, re-query using new information, or explore an interval absent from the retrieved set. Exploration ends when the Reasoner selects $a^{\mathrm{final}}$ or the fixed budget $N$ is exhausted. At finalization, the Reasoner is instructed to rely on the strongest observations available and to prioritize those marked as verified in the ledger.

\subsection{Temporal Evidence Ledger}
\label{sec:ledger}

VESTA retains accumulated observation details in the structured tool memory $\mathcal{M}_t$ and complements it with a temporal evidence ledger that provides an adaptive, compressed, structured evidence view. The ledger reduces the need for the Reasoner to repeatedly reconstruct question-relevant relations from $\mathcal{M}_t$ alone: it is organized around what the observations establish, when they occur, whether they conflict, and whether the current evidence is sufficient, rather than around the sequence of tool invocations.

VESTA first normalizes outputs from all evidence operations into entries $\mathcal{E}_t=\{e_i\}_{i=1}^{n_t}$. Each entry is represented as
\begin{equation}
e_i=(p_i,f_i,\tau_i,q_i,o_i,c_i),
\end{equation}
where $p_i$ identifies the evidence source and interaction step, $f_i$ and $\tau_i$ denote global frame and temporal intervals, $q_i$ is the triggering query or subquestion, $o_i$ is the observation content, and $c_i$ is optional confidence information.

Above these normalized entries, the ledger derives coverage and evidence relations. It merges inspected intervals to estimate temporal coverage; groups neighboring event candidates to distinguish independent occurrences, duplicate descriptions, and unresolved instances; orders numeric observations and state transitions over time; records conflicts explicitly; and links observations to the subquestions or hypotheses they support or refute. This transforms dispersed observations into temporal, coverage, consistency, and hypothesis relations that the Reasoner can consume directly.

Alongside the accumulated tool memory, the ledger uses the evidence-accounting policy $\mu$ and current exploration state to produce a compact snapshot
\begin{equation}
\mathcal{L}_t=\operatorname{Compact}_{\mu}(\mathcal{E}_t)
=\bigl(\widetilde{\mathcal{E}}_t,\mathcal{C}_t,\mathcal{G}_t\bigr),
\end{equation}
where $\widetilde{\mathcal{E}}_t$ preserves question-relevant evidence with temporal location and provenance, $\mathcal{C}_t$ summarizes inspected regions and coverage sufficiency, and $\mathcal{G}_t$ represents events, state transitions, conflicts, verification outcomes, and hypothesis-support relations. Compression prioritizes evidence that distinguishes hypotheses, conflicts with other observations, or still requires verification. It exposes missing evidence and next-step constraints to the Reasoner, thereby guiding subsequent acquisition and presenting verification outcomes for use during finalization.

\section{Experiments}

\subsection{Experimental Setup}

\paragraph{Benchmarks.}
We evaluate on four benchmarks selected for complementary temporal regimes. \textbf{Video-MME-v2} \citep{fu2026videommev2} contains 800 videos and 3,200 eight-choice questions, and scores not only per-question accuracy (Avg) but also Level~1--3 capability levels, Capability Consistency, and Reasoning Coherence across related questions.

\textbf{LongVideoBench} \citep{wu2024longvideobench} contains 3,763 videos and 6,678 multiple-choice questions with durations up to one hour, interleaving frames with subtitles; its \emph{referring reasoning} questions require locating referred contexts in lengthy multimodal inputs before reasoning over local details. We evaluate the long subset of the validation set (564 questions from 188 videos, 15--60 minutes), where locating the right temporal region is the dominant difficulty.

\textbf{EgoSchema} \citep{mangalam2023egoschema} contains more than 5,000 five-choice questions over 250 hours of egocentric video from Ego4D, each posed on a three-minute clip and emphasizing integration of temporally distant events. We use the public 500-question evaluation subset, hereafter \emph{EgoSchema-500}; its compact evidence windows probe whether policy steering preserves performance when long-range search is not the bottleneck.

\textbf{LVBench} \citep{wang2025lvbench} contains 1,549 four-choice questions over 103 videos averaging about 68 minutes (117 hours in total), requiring gathering dispersed evidence and tracking entities across distant intervals.

\paragraph{Ablation subset.}
Because full agent evaluation requires multiple Reasoner and evidence-operation calls, we stratify 400 Video-MME-v2 questions across the benchmark's categories and conduct all mechanism ablations on this fixed subset. Every ablation uses the same questions, videos, evaluation script, model configuration, and interaction budget. Ablations are scored by per-question accuracy (the ``Avg'' metric), not the group-level weighted scores of Video-MME-v2.

\paragraph{Baselines.}
We compare VESTA with five strong MLLMs and the agentic VideoARM baseline on Video-MME-v2. The commercial models are GPT-5 \citep{openai2025gpt5} and Kimi-K2.5 \citep{kimiteam2026kimik25}; the open-source models are Qwen2.5-VL-72B \citep{bai2025qwen2}, LLaVA-Video-72B \citep{zhang2025llavavideo}, and InternVL3.5-241B \citep{wang2025internvl35}. Their results are the official without-subtitle/audio scores reported by Video-MME-v2 \citep{fu2026videommev2}. VideoARM \citep{yin2026videoarm} is our most direct agentic baseline: both systems use OpenAI o3 as the Reasoner, GPT-4o for visual perception, and faster-whisper-large-v3 for speech transcription, although their interaction budgets differ. This comparison therefore controls the principal query-time models but is not a controlled mechanism ablation. On LongVideoBench, EgoSchema, and LVBench, we rerun VideoARM with the same query-time Reasoner, visual-perception, and speech-transcription models as VESTA (replacing its originally reported whisper-1 component with the same local ASR); we mark these reruns VideoARM$^{*}$.
\paragraph{Implementation details.}
VESTA uses OpenAI o3 for the intent router and the Reasoner, GPT-4o for visual evidence operations, and faster-whisper-large-v3 \citep{radford2023robust} for speech transcription. The router receives only the question and candidate answers and neither accesses the video nor participates in verification. We set the maximum number of Reasoner decisions to $N\in\{15,20\}$; because local retrieval decisions count toward $N$ while consuming no remote API tokens, the number of remote-consuming steps remains on the same order as smaller agentic budgets such as VideoARM's. Scene-cache construction, retrieval policies, and remaining hyperparameters are described in Appendix~\ref{app:implementation-details}.
\subsection{Main Results}

\paragraph{Primary comparison on Video-MME-v2.}
Table~\ref{tab:videommev2-results} presents our primary comparison. VESTA improves over the reported VideoARM result on all six metrics, raising average accuracy from 47.6\% to 50.3\% (+2.7). The largest gain appears at Level~1 (+6.6), while Level~2 and Level~3 improve by 0.6 and 2.0 points; Capability Consistency increases by 3.5 points and Reasoning Coherence by 1.4. Because the latter metrics reward consistent success across grouped questions rather than isolated correct answers, their joint improvement suggests that VESTA's gains extend beyond individual evidence hits to more stable evidence use across related reasoning requirements. Attribution to individual mechanisms is provided by the fixed-configuration ablations in Table~\ref{tab:ablation}; the commercial and open-source rows are context only, as they follow the official without-subtitle/audio setting whereas both agentic systems use ASR during exploration.

\begin{table}[t]
\centering
\small
\caption{Results (\%) on Video-MME-v2. Level~1--3, CC, and RC are the benchmark's group-level scores; Avg is per-question accuracy. External MLLM results use the official without-subtitle/audio setting \citep{fu2026videommev2}.}
\label{tab:videommev2-results}
\renewcommand{\arraystretch}{1.08}
\setlength{\tabcolsep}{2pt}
\begin{tabular*}{\columnwidth}{@{\extracolsep{\fill}}lrrrrrr@{}}
\toprule
\textbf{Method} & \textbf{Level 1} & \textbf{Level 2} & \textbf{Level 3} & \textbf{CC} & \textbf{RC} & \textbf{Avg} \\
\midrule
\rowcolor{black!10}
\multicolumn{7}{l}{\textbf{Commercial Models}} \\
GPT-5 & 32.2 & 28.6 & 21.3 & 28.1 & 23.2 & 44.7 \\
Kimi-K2.5 & 30.5 & \textbf{32.8} & \textbf{21.6} & 28.7 & 24.9 & 46.0 \\
\addlinespace[1pt]
\rowcolor{black!10}
\multicolumn{7}{l}{\textbf{Open-source MLLMs}} \\
Qwen2.5-VL-72B & 15.3 & 13.7 & 12.6 & 14.1 & 12.9 & 30.3 \\
LLaVA-Video-72B & 14.8 & 11.1 & 9.5 & 12.3 & 9.5 & 27.3 \\
InternVL3.5-241B & 18.1 & 16.3 & 14.1 & 17.2 & 13.2 & 32.9 \\
\addlinespace[1pt]
\rowcolor{black!10}
\multicolumn{7}{l}{\textbf{Video Agents}} \\
VideoARM & 40.2 & 29.4 & 19.6 & 29.8 & 24.2 & 47.6 \\
\rowcolor{black!18}
\textbf{VESTA (Ours)} & \textbf{46.8} & 30.0 & \textbf{21.6} & \textbf{33.3} & \textbf{25.6} & \textbf{50.3} \\
\bottomrule
\end{tabular*}
\end{table}

\begin{figure}[t]
    \centering
    \includegraphics[width=\columnwidth]{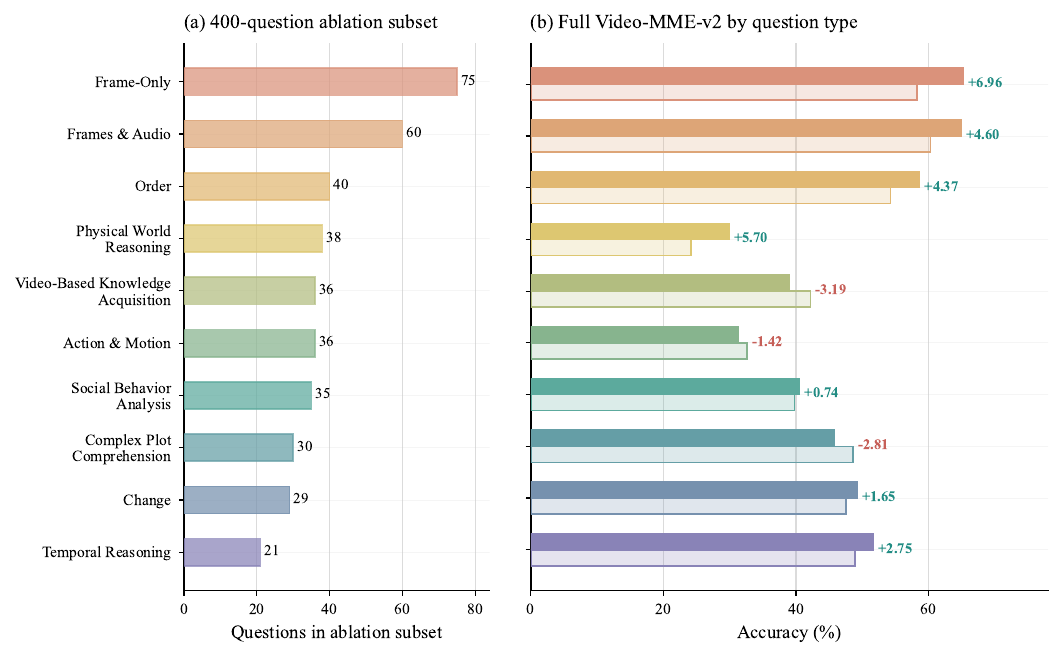}
    \caption{Question-type analysis on Video-MME-v2. (a) Distribution of the fixed 400-question subset used for all mechanism ablations. (b) Accuracy on all 3,200 questions grouped by the benchmark's \texttt{second\_head} annotation. Within each pair, the upper solid bar is VESTA and the lower outlined bar is VideoARM; signed labels report VESTA minus VideoARM in percentage points.}
    \label{fig:videommev2-type-analysis}
\end{figure}

\paragraph{Performance by question type.}
Figure~\ref{fig:videommev2-type-analysis}(a) first verifies that the fixed ablation subset is not skewed toward categories favorable to VESTA: every category's share differs by less than one percentage point from the full benchmark. On all 3,200 questions, Figure~\ref{fig:videommev2-type-analysis}(b) shows that VESTA improves over VideoARM in seven of ten categories, covering 74.6\% of the benchmark, with gains following a coherent evidence-demand pattern. The largest improvement is on Frame-Only questions (+7.0 points), showing that the benefit is not driven solely by speech transcription: focused localization and subsequent visual inspection help the Reasoner recover decisive appearance evidence without repeatedly searching unrelated intervals. Frames \& Audio (+4.6) requires aligning what is said with what is visible at the corresponding time, and Order (+4.4) requires retaining and ordering observations from separated moments---precisely the settings in which temporally grounded evidence operations and the ledger's cross-observation structure should help.

Physical World Reasoning also improves substantially (+5.7), but its absolute accuracy remains below 30\%: better evidence acquisition exposes the relevant objects, actions, and state changes, yet cannot by itself eliminate the downstream reasoning bottleneck. The smaller gains on Temporal Reasoning, Change, and Social Behavior Analysis, and the decreases on Video-Based Knowledge Acquisition, Complex Plot Comprehension, and Action \& Motion, reinforce the same distinction. Overall, the category results support VESTA's claim specifically where questions benefit from locating, aligning, and relating dispersed evidence.

\subsection{Results on Long-Video Benchmarks}
\label{sec:additional-benchmarks}

Table~\ref{tab:longvideobench-shared-model-results} reports results on the LongVideoBench long subset. VESTA improves over the VideoARM$^{*}$ rerun by 6.9 points and also surpasses the strongest previously reported number in the table. The gain is consistent with a benchmark whose main difficulty is not only understanding local content but first finding the right temporal region in a long multimodal stream: policy-steered retrieval reduces the search space before expensive visual inspection, and the ledger keeps the retrieved evidence organized as the reasoning context expands.

\begin{table}[!htbp]
\centering
\small
\setlength{\tabcolsep}{5pt}
\renewcommand{\arraystretch}{1.08}
\caption{Accuracy (\%) on the LongVideoBench 15--60 minute subset. VideoARM$^{*}$ denotes our rerun with the same query-time models as VESTA.}
\label{tab:longvideobench-shared-model-results}
\begin{tabular*}{\columnwidth}{@{\extracolsep{\fill}}lr@{}}
\toprule
\textbf{Method} & \textbf{Accuracy} \\
\midrule
\rowcolor{black!10}
\multicolumn{2}{l}{\textbf{Commercial Models}} \\
Gemini-1.5-Pro & 58.6 \\
Gemini-2.0-Flash & 45.7 \\
GPT-4o & 60.9 \\
OpenAI o3 & 60.6 \\
\addlinespace[1pt]
\rowcolor{black!10}
\multicolumn{2}{l}{\textbf{Video Agents}} \\
DVD & 68.6 \\
VideoARM$^{*}$ & 65.1 \\
\rowcolor{black!18}
\textbf{VESTA (Ours)} & \textbf{72.0} \\
\bottomrule
\end{tabular*}
\end{table}

On EgoSchema (Table~\ref{tab:egoschema-shared-model-results}), VESTA matches VideoARM$^{*}$ at 73.8\%. The tie reflects benchmark structure: each question is posed over a single three-minute clip whose evidence window is compact and temporally well-posed, so retrieval steering has little long-range search to remove---and, importantly, routing does not degrade an already well-matched acquisition regime.

\begin{table}[!htbp]
\centering
\small
\setlength{\tabcolsep}{5pt}
\renewcommand{\arraystretch}{1.08}
\caption{Accuracy (\%) on the public EgoSchema-500 subset.}
\label{tab:egoschema-shared-model-results}
\begin{tabular*}{\columnwidth}{@{\extracolsep{\fill}}lr@{}}
\toprule
\textbf{Method} & \textbf{Accuracy} \\
\midrule
\rowcolor{black!10}
\multicolumn{2}{l}{\textbf{Commercial Models}} \\
Gemini-1.5-Pro & 71.1 \\
Gemini-2.0-Flash & 71.2 \\
GPT-4o & 72.2 \\
OpenAI o3 & 63.2 \\
\addlinespace[1pt]
\rowcolor{black!10}
\multicolumn{2}{l}{\textbf{Video Agents}} \\
VideoAgent & 63.2 \\
VideoTree & 67.0 \\
VideoARM$^{*}$ & \textbf{73.8} \\
\rowcolor{black!18}
\textbf{VESTA (Ours)} & \textbf{73.8} \\
\bottomrule
\end{tabular*}
\end{table}

On LVBench (Table~\ref{tab:lvbench-shared-model-results}), VESTA improves from 67.2\% to 68.7\%. This is consistent with LVBench's requirement that evidence be gathered from distant intervals and maintained coherently across a reasoning trajectory that spans videos averaging over one hour: route-conditioned acquisition and the ledger preserve temporal provenance and organize dispersed evidence as the search unfolds.

\begin{table}[!htbp]
\centering
\small
\setlength{\tabcolsep}{5pt}
\renewcommand{\arraystretch}{1.08}
\caption{Accuracy (\%) on LVBench.}
\label{tab:lvbench-shared-model-results}
\begin{tabular*}{\columnwidth}{@{\extracolsep{\fill}}lr@{}}
\toprule
\textbf{Method} & \textbf{Accuracy} \\
\midrule
\rowcolor{black!10}
\multicolumn{2}{l}{\textbf{Commercial Models}} \\
Gemini-1.5-Pro & 33.1 \\
GPT-4o & 48.9 \\
\addlinespace[1pt]
\rowcolor{black!10}
\multicolumn{2}{l}{\textbf{Open-Source Models}} \\
Qwen2-VL-72B~\citep{wang2024qwen2} & 41.3 \\
AdaReTaKe~\citep{wang2025adaretake} & 53.3 \\
\addlinespace[1pt]
\rowcolor{black!10}
\multicolumn{2}{l}{\textbf{Video Agents}} \\
VideoLucy~\citep{zuo2025videolucy} & 58.8 \\
VideoARM$^{*}$ & 67.2 \\
\rowcolor{black!18}
\textbf{VESTA (Ours)} & \textbf{68.7} \\
\bottomrule
\end{tabular*}
\end{table}

Taken together, the gain grows with the temporal span over which evidence must be located---largest on the 15--60 minute LongVideoBench subset, present on hour-long LVBench videos, and absent-but-harmless on three-minute EgoSchema clips---supporting the claim that VESTA's benefit lies in question-matched evidence acquisition over long timelines rather than in downstream reasoning itself.

\paragraph{Token cost.}
VESTA's cost separates a local, query-independent scene cache from per-question exploration. The cache is built once per video and reused across questions; retrieval against it is local, sub-second, and consumes no remote tokens---which is also why VESTA's nominal step budget ($N\in\{15,20\}$) counts more decisions than agentic baselines whose budgets cover only remote calls \citep{yin2026videoarm}. Since only remote decisions are paid, VESTA's actual token consumption stays in the same range as such baselines rather than scaling with its nominal budget. At question time, only Reasoner decisions and multimodal evidence operations are paid. Budgeting each Reasoner decision at a nominal 8K tokens of context, the reasoning side alone is bounded by $20\times 8{,}000=0.16$M; the measured totals in Appendix~\ref{app:cost-analysis}---which additionally include every remote visual and speech call, where a single vision inspection can itself carry over 10K image tokens---remain on the order of $10^{4}$--$10^{5}$ tokens per question, so these measurements concern cost accounting only and do not affect the accuracy results above. Guidance pays for itself: retrieval-guided trajectories terminate in fewer iterations, whereas unguided exploration burns its budget re-inspecting overlapping intervals. By contrast, dense preprocessing such as DVD pays $\sim$3.98M tokens to ingest a 30-minute video before seeing the question---one to two orders of magnitude more than VESTA's entire exploration, spent largely on regions the question never touches.

\begin{figure}[t]
    \centering
    \includegraphics[width=\columnwidth]{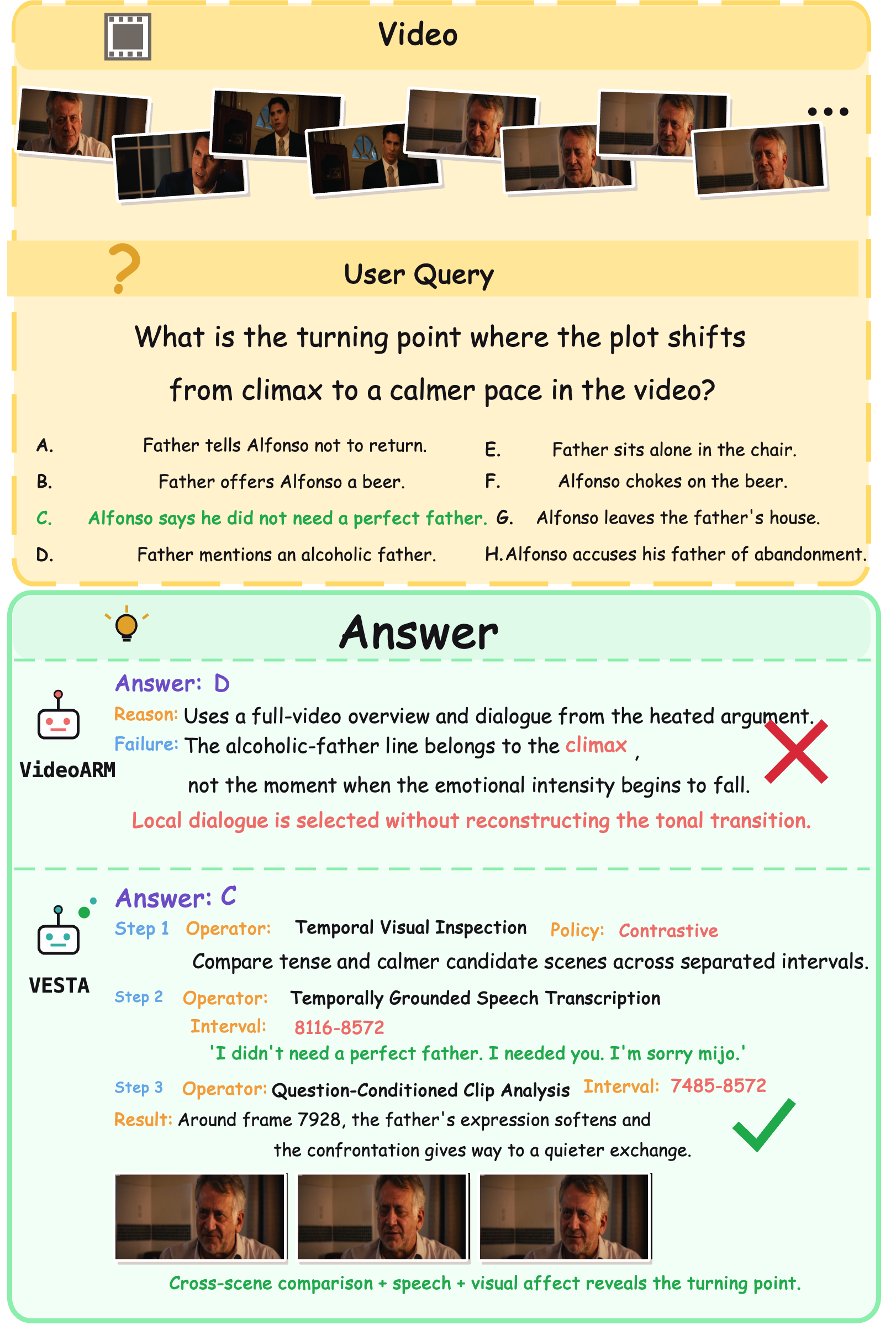}
    \caption{Qualitative comparison on a narrative turning-point question.}
    \label{fig:case-study}
\end{figure}

\subsection{Ablation Study}

\paragraph{Ablation configurations.}
The three mechanisms form a dependency chain rather than three independent switches, so not all on/off combinations are meaningful. The router outputs a pair $(\pi,\mu)$: $\pi$ selects the retrieval strategy, while $\mu$ is an \emph{evidence-accounting policy} that tells the Temporal Evidence Ledger which evidence views to maintain and prioritize under compression. The ledger therefore has nothing to follow if routing is disabled: the two combinations ``routing off, ledger on'' are undefined. Symmetrically, if both retrieval and the ledger are disabled, $(\pi,\mu)$ has no consumer and the configuration reduces exactly to the plain agent without any mechanism, so it adds no information beyond that baseline. This leaves five meaningful configurations, all reported in Table~\ref{tab:ablation}. In the table, ``Routing~--'' fixes the retrieval strategy to a single question-level query with standard top-$K$ ranking and leaves the ledger unconfigured; ``Retrieval~--'' removes the shared scene index so the Reasoner must choose inspection intervals on its own; ``Ledger~--'' lets the Reasoner keep the accumulated structured tool memory but hides the ledger's consolidated evidence view.

We evaluate these configurations on the fixed 400-question Video-MME-v2 subset under identical model, input, evaluation, and interaction-budget settings. Table~\ref{tab:ablation} uses the full VESTA configuration as the reference, which outperforms the variant without routing, retrieval, or ledger mechanisms by 15.8 points. Retrieval alone improves over the no-mechanism variant by 9.3 points, establishing the value of evidence localization. Adding routing on top of retrieval contributes a further 3.0-point gain, showing that question-conditioned acquisition helps beyond generic retrieval. Adding the Temporal Evidence Ledger to route-conditioned retrieval yields another 3.5-point gain, demonstrating that structured evidence consolidation remains beneficial after policy-steered retrieval is present. Conversely, removing retrieval while retaining routing and the ledger produces a 10.0-point drop from the full reference, confirming that evidence accounting cannot substitute for acquiring relevant candidates. Because these components interact, the differences should be interpreted as controlled conditional effects rather than context-independent additive contributions.

\begin{table}[!htbp]
\centering
\small
\setlength{\tabcolsep}{5pt}
\renewcommand{\arraystretch}{1.08}
\caption{Mechanism ablations on the Video-MME-v2 subset. The ledger is configured by the router's policy $\mu$, so disabling routing while keeping the ledger is undefined; the five meaningful configurations are reported in full.}
\label{tab:ablation}
\begin{tabular*}{\columnwidth}{@{\extracolsep{\fill}}cccc@{}}
\toprule
\textbf{Routing} & \textbf{Retrieval} & \textbf{Ledger} &
\shortstack[c]{\textbf{Accuracy}\\[-1pt]{\scriptsize vs. Full}} \\
\midrule
-- & -- & -- &
\shortstack[c]{37.5\\[-1pt]{\scriptsize\color{black!55}$\Delta\ -15.8$}} \\
\checkmark & -- & \checkmark &
\shortstack[c]{43.3\\[-1pt]{\scriptsize\color{black!55}$\Delta\ -10.0$}} \\
-- & \checkmark & -- &
\shortstack[c]{46.8\\[-1pt]{\scriptsize\color{black!55}$\Delta\ -6.5$}} \\
\checkmark & \checkmark & -- &
\shortstack[c]{49.8\\[-1pt]{\scriptsize\color{black!55}$\Delta\ -3.5$}} \\
\rowcolor{black!18}
\checkmark & \checkmark & \checkmark &
\shortstack[c]{\textbf{53.3}} \\
\bottomrule
\end{tabular*}
\end{table}

\section{Conclusion}

We introduced VESTA, a training-free long-video agent that addresses a structural mismatch between question-specific evidence requirements and the uniform acquisition behavior common in existing agents. VESTA organizes exploration as a route-conditioned acquire--verify--consolidate loop: an intent router selects focused, recall, or contrastive acquisition, retrieved scenes remain provisional references that preserve Reasoner autonomy, and a temporal evidence ledger maintains an adaptive, compressed view of observations, verification outcomes, coverage, conflicts, and unresolved requirements. Across Video-MME-v2 and three long-video benchmarks, gains scale with the temporal span over which evidence must be located, and controlled ablations attribute complementary benefits to policy-steered retrieval and structured evidence accounting. The current instantiation uses a fixed policy set and proprietary models for query-time reasoning and visual verification; extending the policy space and evaluating fully open-model configurations remain future work.
\clearpage
\bibliography{aaai2027}

\clearpage
\appendix

\section{Additional Experiments}
\label{app:additional-experiments}

\subsection{Ablation Subset Composition}
\label{app:ablation-subset}

\begin{figure}[!htbp]
    \centering
    \includegraphics[width=\columnwidth]{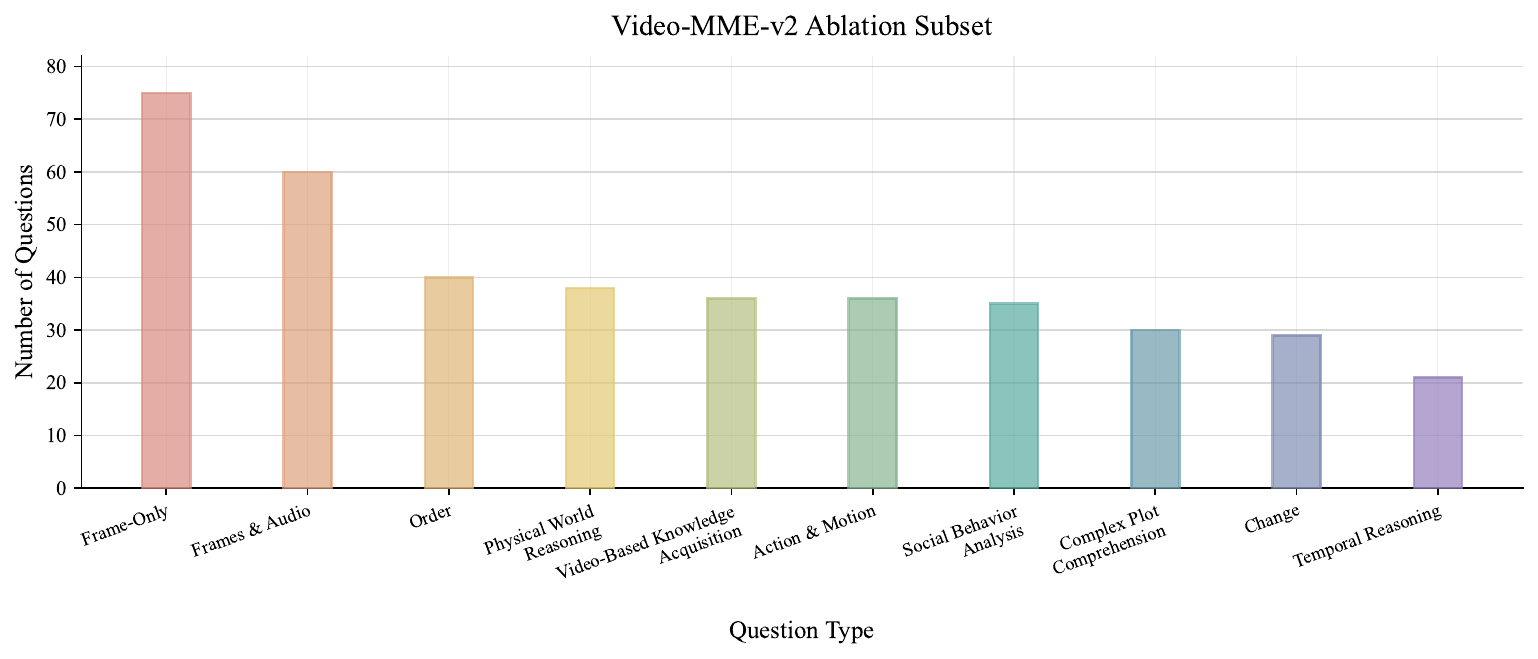}
    \caption{Question-type distribution of the fixed 400-question Video-MME-v2 ablation subset.}
    \label{fig:ablation-subset}
\end{figure}

Figure~\ref{fig:ablation-subset} shows the question-type distribution of the fixed 400-question Video-MME-v2 subset used for all mechanism ablations. The subset covers all ten official question categories and closely preserves the full benchmark's composition: the share of every category differs by less than one percentage point from its share among all 3,200 questions. This alignment reduces the risk that the ablation conclusions are driven by over-representing a small set of question types.

\subsection{Routing and Budget Trends}
\label{app:routing-budget-trends}

\begin{figure*}[!htbp]
    \centering
    \includegraphics[width=0.48\textwidth]{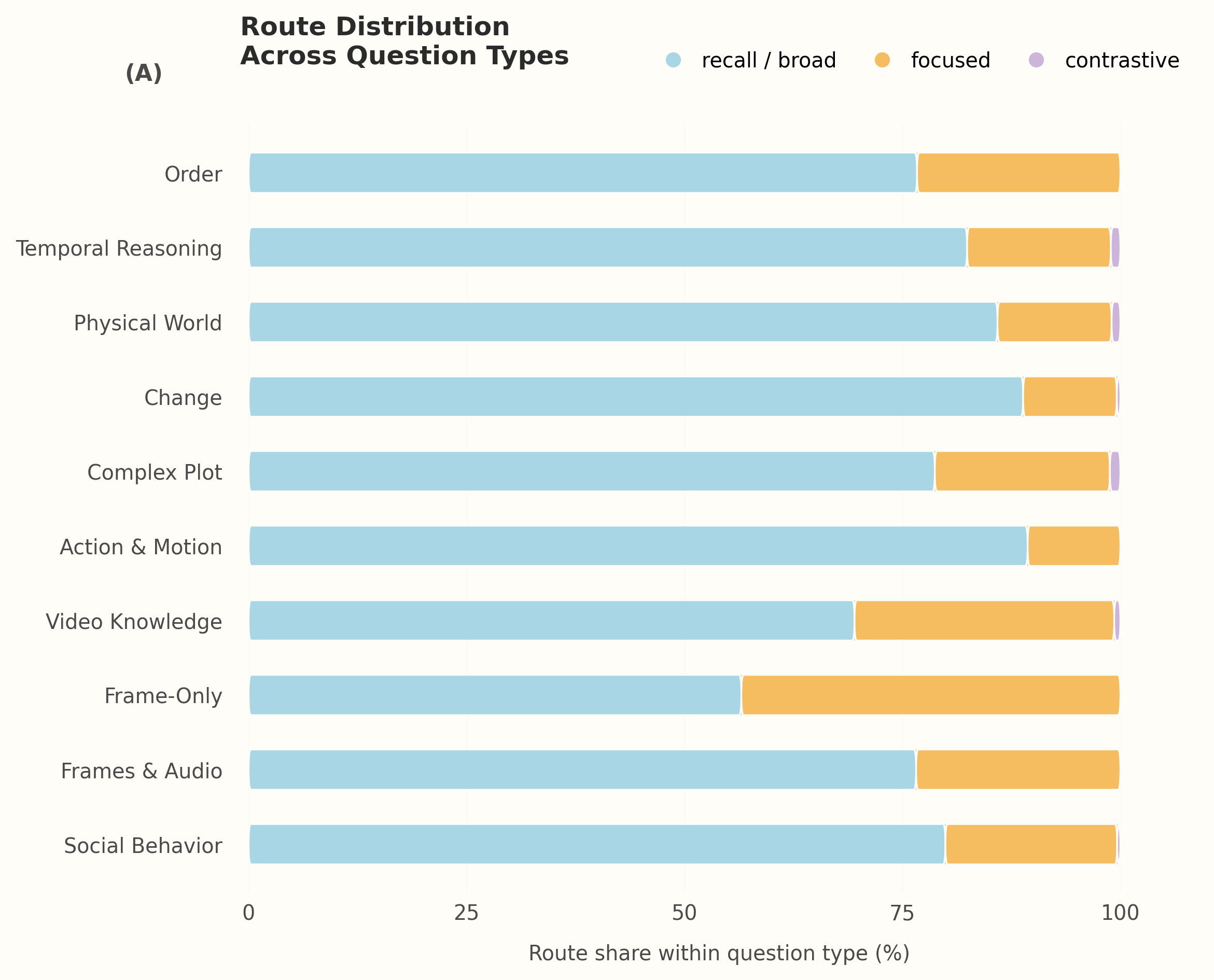}
    \hfill
    \includegraphics[width=0.48\textwidth]{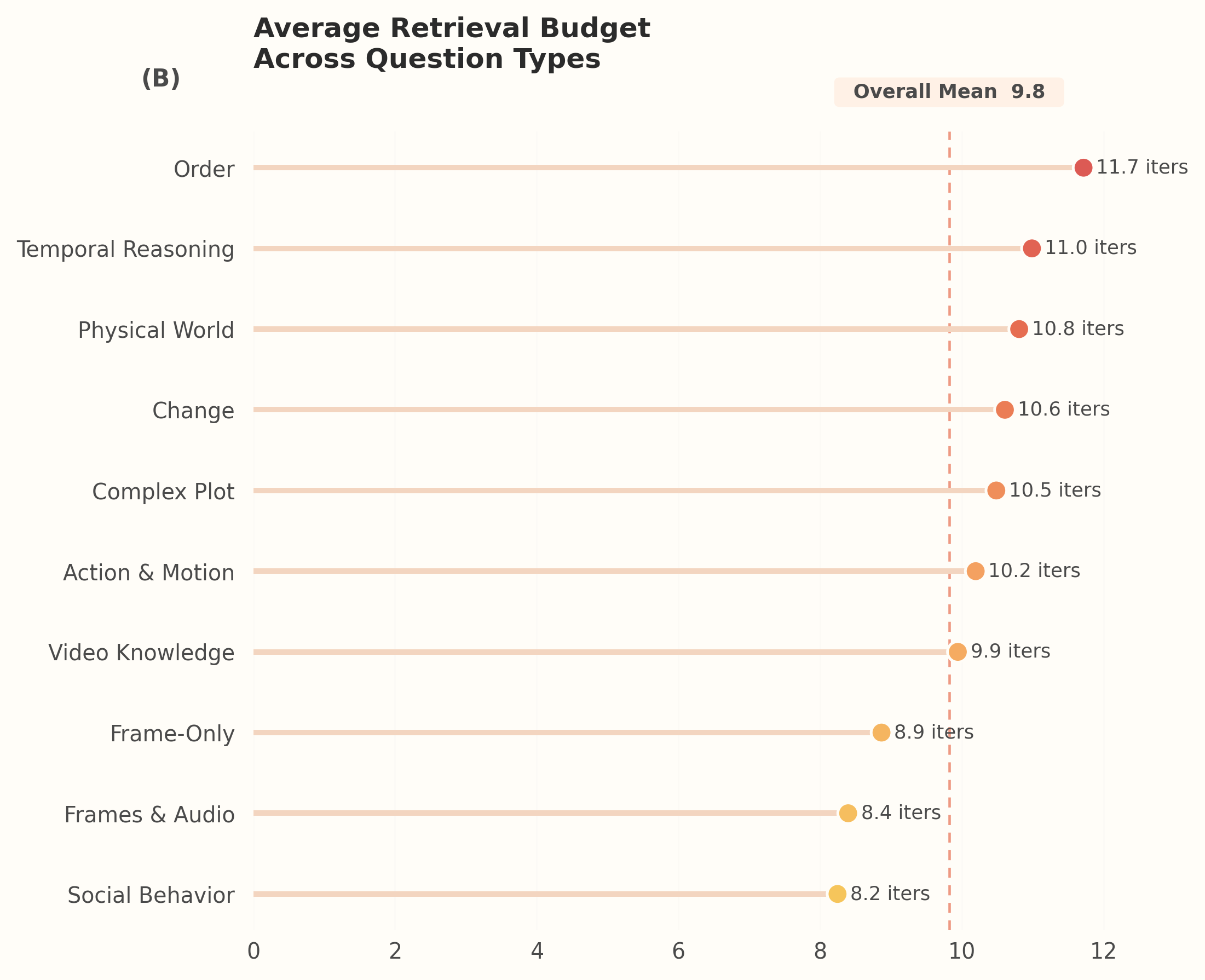}
    \caption{Routing and interaction-budget trends across Video-MME-v2 question categories. The left panel shows the distribution of focused, broad, and contrastive acquisition routes; the right panel shows the corresponding mean interaction budget.}
    \label{fig:second-head-route-budget}
\end{figure*}

Figure~\ref{fig:second-head-route-budget} summarizes how VESTA's acquisition routes and interaction budgets vary across Video-MME-v2 question categories. The route distribution is broadly consistent with the intended localization--coverage trade-off. Categories that require repeated-event coverage, temporal comparison, or evidence across separated regions predominantly use broad, recall-oriented exploration. Frame-Only questions exhibit a more balanced mixture of focused and broad routes, reflecting the coexistence of localized visual-recognition questions and questions that require comparison across multiple frames. Frames-and-Audio and Social Behavior Analysis tend toward more localized budgets, consistent with resolving multimodal or interactional evidence within a limited temporal context. Complex Plot Comprehension shows a comparatively larger role for focused and selective contrastive exploration, suggesting that some narrative questions are better served by testing specific interpretations than by uniformly expanding temporal coverage.

The budget trend follows the same evidence-demand structure. Order, temporal, physical-world, and other globally grounded categories tend to require longer interaction trajectories, whereas localized visual and social questions more often terminate after shorter exploration. Contrastive routing remains selective rather than becoming a default response to difficulty. Overall, these trends indicate that the router adapts both search breadth and evidence-accounting effort to the anticipated structure of the question. At the same time, larger budgets do not guarantee higher accuracy: additional evidence improves availability and organization, but does not eliminate downstream temporal or semantic reasoning bottlenecks.

\subsection{Additional Benchmark Evaluations}
\label{app:additional-benchmarks}

Dataset descriptions, the shared-model evaluation configuration, and results on the LongVideoBench long subset, EgoSchema-500, and LVBench are reported in the main paper (Section~\ref{sec:additional-benchmarks}). Appendix~\ref{app:ledger-case-studies} additionally provides two LVBench ledger case studies as mechanism examples.

\subsection{Qualitative Case Studies}
\label{app:qualitative-analysis}

Figure~\ref{fig:case-study} in the main paper illustrates a narrative-turning-point question. The baseline selects dialogue from the climax but does not reconstruct the transition toward a calmer exchange. VESTA routes the question to contrastive acquisition, compares candidate scenes from separated intervals, and combines Temporally Grounded Speech Transcription with Question-Conditioned Clip Analysis to verify the tonal transition. This case shows how question-conditioned acquisition and cross-scene evidence consolidation can correct an initially plausible but locally grounded answer.

In a separate fixed-$N=10$ diagnostic run, we examine a focused question asking for the complete name on a circular golden sign in an 844.8-second video. Without retrieval guidance, the baseline spends all ten iterations repeatedly inspecting overlapping regions around 15--133 seconds, while the target appears near 540 seconds, and answers without readable textual evidence. VESTA routes the question to focused acquisition and retrieves frames 32,413--32,743 (approximately 540--546 seconds). Temporal Visual Inspection confirms the storefront, and Question-Conditioned Clip Analysis narrows the evidence to frames 32,413--32,435 and reads ``TOOTHSOME CHOCOLATE EMPORIUM.'' VESTA therefore answers correctly in its third iteration. Retrieval does not replace multimodal evidence operations in this case; it provides an evidence-based location that converts unguided interval guessing into candidate localization, local observation, and fine-grained verification.

\subsection{Evidence-Acquisition Cost Analysis}
\label{app:cost-analysis}

Accuracy alone does not characterize where a long-video agent spends its budget, so we use Video-MME-v2 UID~015-1 as a single-question diagnostic. The three configurations all answer correctly, but they distribute remote effort differently across reasoning, retrieval, and visual inspection.

Table~\ref{tab:cost-case} shows that the main cost differences come from the shape of the downstream Reasoner and visual-analysis calls, not from answer correctness itself. The configuration with Routing and the Temporal Evidence Ledger but without Retrieval spends the most, Retrieval only spends the least, and Routing with Retrieval but without the Temporal Evidence Ledger lies in between. This pattern suggests that Retrieval should not be interpreted as a standalone cost source; its cost effect depends on how the retrieved candidates are converted into verified observations and maintained in the reasoning context. 

\begin{table}[!htbp]
\centering
\scriptsize
\caption{Provider usage for Video-MME-v2 UID~015-1. All configurations answer correctly.}
\label{tab:cost-case}
\setlength{\tabcolsep}{2pt}
\renewcommand{\arraystretch}{1.08}
\begin{tabular*}{\columnwidth}{@{\extracolsep{\fill}}lrrrrrr@{}}
\toprule
\textbf{Configuration} & \shortstack[r]{\textbf{Reasoner}\\\textbf{iter.}} & \shortstack[r]{\textbf{Remote}\\\textbf{req.}} & \shortstack[r]{\textbf{Failed}\\\textbf{req.}} & \shortstack[r]{\textbf{Input}\\\textbf{tokens}} & \shortstack[r]{\textbf{Output}\\\textbf{tokens}} & \shortstack[r]{\textbf{Total}\\\textbf{tokens}} \\
\midrule
\shortstack[l]{Routing + Temporal\\Evidence Ledger, no Retrieval} & 4 & 8 & 0 & 80{,}766 & 1{,}832 & 82{,}598 \\
\shortstack[l]{Routing + Retrieval, no\\Temporal Evidence Ledger} & 7 & 13 & 1 & 57{,}935 & 2{,}651 & 60{,}586 \\
Retrieval only & 5 & 8 & 0 & 25{,}136 & 1{,}766 & 26{,}902 \\
\bottomrule
\end{tabular*}
\end{table}

Table~\ref{tab:cost-operations} separates local Semantic Scene Retrieval from the remote trajectory. Semantic Scene Retrieval adds little direct latency; the meaningful cost difference comes from how it changes the number of Temporal Visual Inspection calls and the amount of evidence that must be consolidated. In that sense, Retrieval and the Temporal Evidence Ledger affect cost indirectly by shaping the search trajectory rather than by making the retrieval operation itself expensive.

\begin{table}[!htbp]
\centering
\scriptsize
\caption{Evidence-operation usage for Video-MME-v2 UID~015-1. SSR, TVI, and QCCA denote Semantic Scene Retrieval, Temporal Visual Inspection, and Question-Conditioned Clip Analysis, respectively.}
\label{tab:cost-operations}
\setlength{\tabcolsep}{1pt}
\renewcommand{\arraystretch}{1.12}
\begin{tabular*}{\columnwidth}{@{\extracolsep{\fill}}p{0.34\columnwidth}rrrrr@{}}
\toprule
\textbf{Configuration} & \shortstack[r]{\textbf{SSR}\\\textbf{calls}} & \shortstack[r]{\textbf{TVI}\\\textbf{calls}} & \shortstack[r]{\textbf{QCCA}\\\textbf{calls}} & \textbf{Frames} & \shortstack[r]{\textbf{SSR}\\\textbf{time (s)}} \\
\midrule
Routing + Temporal Evidence Ledger, no Retrieval & 0 & 1 & 2 & 90 & -- \\
Routing + Retrieval, no Temporal Evidence Ledger & 2 & 4 & 0 & 210 & 0.136 \\
Retrieval only & 1 & 3 & 0 & 90 & 0.077 \\
\bottomrule
\end{tabular*}
\end{table}

For a rough baseline comparison, DVD's fixed preprocessing pipeline imposes an order-of-magnitude larger visual-input burden than the on-demand VESTA trajectories. The relevant point is the scale separation: VESTA pays for evidence only when the question requires it, rather than front-loading a dense scan of the whole video.

\subsection{Ledger Case Studies}
\label{app:ledger-case-studies}

The following two LVBench examples illustrate how the ledger organizes retrieval references, verified observations, temporal coverage, and task-specific evidence relations. They are presented as mechanism examples rather than as additional benchmark results.

\begin{table*}[!htbp]
\centering
\scriptsize
\caption{Representative LVBench ledger entries. Entry groups retain the ledger's provenance dimensions while aggregating repeated or closely related records for readability; retrieval remains provisional and Reasoner synthesis is shown separately.}
\label{tab:ledger-case-studies}
\setlength{\tabcolsep}{3pt}
\renewcommand{\arraystretch}{1.1}
\begin{tabular*}{\textwidth}{@{\extracolsep{\fill}}p{0.12\textwidth}p{0.15\textwidth}p{0.18\textwidth}p{0.34\textwidth}p{0.16\textwidth}@{}}
\toprule
\textbf{Evidence group} & \textbf{Evidence Operation / Stage} & \textbf{Temporal Provenance} & \textbf{Evidence Content} & \textbf{Ledger Representation} \\
\midrule
\rowcolor{black!10}
\multicolumn{5}{l}{\textbf{UID 1964}} \\
E2--E7 & Semantic Scene Retrieval & Multiple intervals & Query about hosts without chest badges; several candidate regions are returned. & Provisional references \\
E8--E9 & Temporal Visual Inspection & [21996, 22150] & A formal ceremony with front-row hosts is observed after narrowing the retrieved interval. & Verified visual evidence \\
E10--E11 & Question-Conditioned Clip Analysis & [21996, 22150] & Two analyses return the count four on the same narrowed interval. & Repeated numeric observations grouped into one temporal cluster; no conflict \\
\rowcolor{black!18}
Derived & Reasoner Synthesis & Cross-observation & The repeated count on the verified interval supports answer A. & Derived from the numeric evidence cluster \\
\midrule
\rowcolor{black!10}
\multicolumn{5}{l}{\textbf{UID 1426}} \\
E1--E6 & Semantic Scene Retrieval & Multiple intervals & Query about people preparing food; several temporally separated candidate regions are returned. & Provisional references \\
E7--E9 & Temporal Visual Inspection & Three related intervals & People prepare food in outdoor and communal settings across multiple inspected regions. & Verified observations merged into a compact coverage view \\
Summary & Temporal Evidence Ledger & Three merged ranges & The question requires a visual commonality rather than counting, event, or audio-specific structure. & Lightweight task-conditioned evidence view \\
\rowcolor{black!18}
Derived & Reasoner Synthesis & Cross-observation & The shared visual attribute across inspected cooking scenes supports answer A. & Derived from multi-interval visual evidence \\
\bottomrule
\end{tabular*}
\end{table*}

These cases illustrate two complementary uses of the ledger. In UID~1964, it links provisional retrieval references to a narrowed visual interval and consolidates repeated numeric observations without confusing temporal deduplication with the answer value. In UID~1426, it organizes several verified intervals into a compact coverage view while retaining a lightweight accounting structure appropriate to a non-counting question. In both cases, retrieval identifies where evidence may be found, perceptual operations establish observations, and the ledger preserves their temporal provenance and task-relevant relations for the Reasoner.

\section{Algorithmic Details}
\label{app:algorithmic-details}

\subsection{Implementation Details}
\label{app:implementation-details}

We construct the query-independent scene cache using PySceneDetect with threshold 27.0 and 5--60 second scenes, sampling at most four frames per scene and aligning overlapping ASR. Qwen3-VL-Embedding \citep{li2026qwen3vlembedding} encodes each visual--speech scene and text query into a 2,048-dimensional space, and NanoVectorDB stores the scene representations and temporal metadata. Cache construction, encoding, and retrieval run locally and incur no closed-model API calls.

All retrieval policies use an initial budget of $K=6$: focused retrieval returns the top six scenes; recall retrieval expands to at most 15 candidates, favors scenes separated by at least 20 seconds, and backfills when needed; contrastive retrieval distributes the six-scene budget across at most six candidate hypotheses and temporally de-duplicates the merged results. Visual inspection and local clip analysis process at most 150 and 50 frames per call, respectively.

This section gives an implementation-oriented description of VESTA's persistent state, route-conditioned agent loop, policy-steered retrieval, and ledger update. All temporal locations use global video coordinates. Retrieval outputs remain provisional references; only observations produced by multimodal evidence operations provide perceptual content for reasoning and finalization.

\subsection{Persistent State and Ledger Snapshot}

At interaction step $t$, VESTA maintains accumulated structured tool memory $\mathcal{M}_t$ together with the Temporal Evidence Ledger $\mathcal{L}_t$. The former preserves detailed operation outputs. The latter normalizes observations as entries $e_i=(p_i,f_i,\tau_i,q_i,o_i,c_i)$ and exposes a compact relational view over temporal coverage, event and state relations, conflicts, verification outcomes, and unresolved requirements. Compaction changes only the snapshot presented to the Reasoner; it does not remove the detailed observations retained in $\mathcal{M}_t$.

\subsection{Intent Routing and Policy Projection}
\label{app:intent-routing}

Before video exploration, VESTA invokes the intent router once using only the question $Q$ and optional candidate answers $C$. The router does not access video content, subtitles, benchmark identity, question-type annotations, or the ground-truth answer. Its role is therefore limited to determining how evidence should be acquired and organized; it neither performs video perception nor predicts the answer.

The router combines a text-based intent estimate with a deterministic prior derived from observable linguistic cues in the question and candidate answers. Let
\begin{equation}
\rho=\mathcal{R}_{\mathrm{model}}(Q,C), \qquad
b=\mathcal{R}_{\mathrm{prior}}(Q,C),
\end{equation}
where $\rho$ captures the model-estimated evidence demand and $b$ encodes stable requirements associated with localized recognition, repeated-event coverage, ordering, quantitative comparison, and cross-modal reasoning. The composite router used in the main paper is realized as
\begin{equation}
(\pi,\mu)=
\operatorname{Project}\!\left(
\operatorname{Merge}(b,\rho)
\right).
\end{equation}
The merge preserves explicit evidence requirements while allowing the model estimate to refine ambiguous semantic or hypothesis-discriminative cases. If the model estimate is unavailable or insufficiently confident, VESTA retains the deterministic prior and continues with the corresponding conservative policy rather than terminating exploration.

\begin{table}[!htbp]
\centering
\small
\caption{Projection from evidence demand to acquisition and accounting policies.}
\label{tab:intent-policy-projection}
\setlength{\tabcolsep}{3pt}
\renewcommand{\arraystretch}{1.08}
\begin{tabular*}{\columnwidth}{@{\extracolsep{\fill}}p{0.34\columnwidth}p{0.18\columnwidth}p{0.38\columnwidth}@{}}
\toprule
\textbf{Evidence demand} & \textbf{$\pi$} & \textbf{Emphasis of $\mu$} \\
\midrule
Localized recognition or description & Focused & Compact local evidence \\
Repeated events, counting, or ordering & Recall & Temporal coverage and event relations \\
Competing answers or interpretations & Contrastive & Hypothesis support and conflicts \\
\bottomrule
\end{tabular*}
\end{table}

The projected acquisition policy $\pi$ determines how the initial temporal references are constructed, while the accounting policy $\mu$ determines which coverage, relational, and unresolved-evidence information is emphasized in the ledger snapshot. Both policies are exposed to the Reasoner as guidance, but neither prescribes a fixed reasoning procedure or answer. Contrastive hypothesis decomposition is used for initial candidate construction; subsequent retrieval calls are formulated by the Reasoner according to its evolving evidence needs.

\subsection{Route-Conditioned Agent Loop}

Algorithm~\ref{alg:vesta-agent} summarizes the complete inference loop. The intent router is invoked once before exploration and does not access the video. The selected policy structures the initial references and ledger view, while the Reasoner retains control over subsequent evidence operations and may re-query or inspect regions outside the initial candidate set.

\begin{algorithm}[!htbp]
\caption{Route-conditioned VESTA inference}
\label{alg:vesta-agent}
\begin{algorithmic}[1]
\Require video $V$, question $Q$, optional choices $C$, budget $N$
\Ensure answer $A$
\State $(\pi,\mu) \gets \mathcal{R}(Q,C)$
\State $\mathcal{R}_{\pi} \gets \Call{PolicyRetrieve}{Q,C,\pi}$
\State initialize structured tool memory $\mathcal{M}_0$ and ledger state $\mathcal{E}_0$
\For{$t=0,\ldots,N-1$}
    \State $\mathcal{L}_t \gets \Call{CompactLedger}{\mathcal{E}_t,\mu}$
    \State $H_t \gets (Q,\pi,\mathcal{R}_{\pi},\mathcal{L}_t,\mathcal{M}_t)$
    \State $a_t \gets \Call{ReasonerSelect}{H_t}$
    \If{$a_t=a^{\mathrm{final}}$}
        \State \Return $\Call{Finalize}{H_t}$
    \EndIf
    \State $o_t \gets \Call{ExecuteEvidenceOperation}{a_t,V}$
    \State $\mathcal{M}_{t+1} \gets \Call{UpdateToolMemory}{\mathcal{M}_t,o_t}$
    \State $\mathcal{E}_{t+1} \gets \Call{UpdateLedger}{\mathcal{E}_t,o_t,a_t}$
\EndFor
\State $\mathcal{L}_N \gets \Call{CompactLedger}{\mathcal{E}_N,\mu}$
\State \Return $\Call{Finalize}{Q,\pi,\mathcal{R}_{\pi},\mathcal{L}_N,\mathcal{M}_N}$
\end{algorithmic}
\end{algorithm}

\begin{algorithm}[t]
\caption{Policy-steered retrieval}
\label{alg:policy-retrieval}
\begin{algorithmic}[1]
\Require question $Q$, optional choices $C$, policy $\pi$, scene index $\mathcal{S}$, budget $K$
\Ensure provisional reference set $\mathcal{R}_{\pi}$
\If{$\pi=\mathit{focused}$}
    \State $\mathcal{Z} \gets \Call{Search}{Q,\mathcal{S}}$
    \State $\mathcal{R}_{\pi} \gets \Call{TopK}{\mathcal{Z},K}$
\ElsIf{$\pi=\mathit{recall}$}
    \State $\mathcal{Z} \gets \Call{Search}{Q,\mathcal{S}}$
    \State $\mathcal{P} \gets \Call{TopK}{\mathcal{Z},\min(4K,15)}$
    \State $\mathcal{R}_{\pi} \gets \Call{TemporalDisperse}{\mathcal{P},K}$
    \If{$|\mathcal{R}_{\pi}|<K$}
        \State backfill deferred candidates from $\mathcal{P}$ by relevance
    \EndIf
\Else
    \State $\mathcal{H} \gets \Call{ParseHypotheses}{C}$
    \If{$\mathcal{H}=\varnothing$}
        \State $\mathcal{Z} \gets \Call{Search}{Q,\mathcal{S}}$
        \State $\mathcal{R}_{\pi} \gets \Call{TopK}{\mathcal{Z},K}$
    \Else
        \State $\mathcal{R}_{\pi} \gets \varnothing$
        \ForAll{$h_j\in\mathcal{H}$}
            \State $q_j \gets \Call{ComposeQuery}{Q,h_j}$
            \State $\mathcal{Z}_j \gets \Call{Search}{q_j,\mathcal{S}}$
            \State $\mathcal{R}_j \gets \Call{TopK}{\mathcal{Z}_j,k_j}$
            \State $\mathcal{R}_{\pi} \gets \mathcal{R}_{\pi}\cup\mathcal{R}_j$
        \EndFor
        \State $\mathcal{R}_{\pi} \gets \Call{TemporalDeduplicate}{\mathcal{R}_{\pi}}$
        \State truncate $\mathcal{R}_{\pi}$ to at most $K$ references
    \EndIf
\EndIf
\State mark every $r\in\mathcal{R}_{\pi}$ as an unverified provisional reference
\State \Return $\mathcal{R}_{\pi}$
\end{algorithmic}
\end{algorithm}

\subsection{Evidence Operation Workflows}
\label{app:evidence-operation-workflows}

VESTA provides four complementary evidence operations that support localization, perception, multimodal alignment, and targeted verification. Retrieval returns provisional references, while the remaining operations convert selected temporal regions into observations. These observations are retained in structured tool memory and organized by the Temporal Evidence Ledger for subsequent reasoning.

\paragraph{Semantic Scene Retrieval.}
This operation locates video regions that are semantically relevant to the current evidence need. It searches the shared visual--speech scene index and returns ranked temporal references, allowing the Reasoner to begin from plausible locations rather than inspect the timeline uniformly. The returned regions remain unverified until a perceptual operation examines their content.

\paragraph{Temporal Visual Inspection.}
This operation examines selected temporal intervals and produces concise, temporally localized descriptions of their visual content. It serves as the general-purpose visual observation mechanism for identifying objects, actions, settings, and state changes, and can verify or reject locations proposed by retrieval.

\paragraph{Temporally Grounded Speech Transcription.}
This operation recovers speech from selected intervals and aligns each utterance with the global video timeline. The resulting observations allow the Reasoner to connect dialogue, narration, and other spoken information with visual events occurring at the same or nearby times.

\paragraph{Question-Conditioned Clip Analysis.}
This operation performs fine-grained analysis of a local interval in response to a specific unresolved evidence requirement. Rather than producing a general scene description, it answers a targeted sub-question about the selected clip and records the result together with its temporal provenance and confidence, supporting focused verification and conflict resolution.

Answer Finalization is not an evidence operation; it is the terminal Reasoner decision made from the accumulated observations and their structured evidence view.

\subsection{Policy-Steered Retrieval}
Algorithm~\ref{alg:policy-retrieval} makes explicit that all three policies share one scene index and differ only in query construction and candidate-set formation. Recall retrieval broadens temporal coverage within a bounded relevance pool; it does not guarantee recovery of every relevant occurrence. Contrastive retrieval uses parsed answer choices as hypotheses when available and keeps the merged candidate scale bounded.

Three implementation details are worth noting. First, every branch begins with the same $\Call{Search}{\cdot,\mathcal{S}}$ primitive, so the policy affects only how the ranking is consumed, never the index or similarity computation; this guarantees that any accuracy difference in the retrieval-policy ablations is attributable to candidate-set structure rather than to a change of retrieval basis. Second, all candidate sets are bounded: focused and contrastive retrieval return at most $K$ scenes, and recall retrieval never inspects more than $\min(4K,15)$ candidates before dispersion, so the per-question retrieval cost of any policy differs from the others by at most a constant factor and remains local. Third, the fallback paths keep the algorithm total: when the candidate set cannot be parsed into hypotheses, contrastive retrieval degrades to focused retrieval; when too few scenes satisfy the temporal-separation constraint, recall retrieval backfills from the deferred pool. In both cases the Reasoner still receives a well-formed provisional reference set and retains the freedom to re-query or explore outside it.

\subsection{Ledger Update and Compaction}

\begin{algorithm}[t]
\caption{Temporal Evidence Ledger update and compaction}
\label{alg:ledger-update}
\begin{algorithmic}[1]
\Require persistent ledger state $\mathcal{E}_t$, observation $o_t$, operation $a_t$, policy $\mu$
\Ensure updated state $\mathcal{E}_{t+1}$ and compact snapshot $\mathcal{L}_{t+1}$
\State $X_t \gets \Call{NormalizeOperationOutput}{o_t,a_t}$
\ForAll{$x\in X_t$}
    \If{$x$ contains a valid observation}
        \State $e \gets \Call{MakeEntry}{x.\mathit{source},x.\mathit{frames},x.\mathit{time}}$
        \State attach query, content, and confidence fields from $x$ to $e$
        \State append $e$ to $\mathcal{E}_t$
        \State update temporal coverage and provenance relations
        \State update event, state-transition, conflict, and hypothesis relations when supported
    \EndIf
\EndFor
\State $\mathcal{E}_{t+1} \gets \mathcal{E}_t$
\State $\widetilde{\mathcal{E}}_{t+1} \gets \Call{SelectRelevantEntries}{\mathcal{E}_{t+1},\mu}$
\State $\mathcal{C}_{t+1} \gets \Call{SummarizeCoverage}{\mathcal{E}_{t+1},\mu}$
\State $\mathcal{G}_{t+1} \gets \Call{SummarizeRelationsAndGaps}{\mathcal{E}_{t+1},\mu}$
\State $\mathcal{L}_{t+1} \gets (\widetilde{\mathcal{E}}_{t+1},\mathcal{C}_{t+1},\mathcal{G}_{t+1})$
\State \Return $(\mathcal{E}_{t+1},\mathcal{L}_{t+1})$
\end{algorithmic}
\end{algorithm}

After an evidence operation returns, VESTA normalizes the result into one or more observation entries, appends the detailed output to structured tool memory, and updates the ledger relations. Algorithm~\ref{alg:ledger-update} separates this persistent update from policy-conditioned compaction. The snapshot emphasizes question-relevant evidence, coverage gaps, conflicts, and observations that still require verification, while finalization remains a Reasoner decision rather than a programmatic verification gate.

The two phases play complementary roles. The persistent update (lines 2--8) is monotone and policy-free: every valid observation is appended to $\mathcal{E}_t$ with its provenance, and relation updates are applied only ``when supported,'' so no observation is ever discarded or reinterpreted by the accounting policy. Compaction (lines 10--14), by contrast, is a pure view computation over $\mathcal{E}_{t+1}$: $\widetilde{\mathcal{E}}_{t+1}$ selects which entries the Reasoner sees, while $\mathcal{C}_{t+1}$ and $\mathcal{G}_{t+1}$ summarize coverage and relations. Because the snapshot is recomputed from the persistent state at every step, the Reasoner can always recover detail from $\mathcal{M}_t$, and a change of emphasis under $\mu$ cannot lose evidence---it can only change what is surfaced next. This separation is what allows the ablation that disables the ledger to retain the full structured tool memory without altering any observation.


\end{document}